\documentclass[fleqn,10pt]{wlscirep}
\makeatletter
\let\@@showhyphens\showhyphens
\renewcommand{\showhyphens}[1]{%
  \typeout{Hyphenation for: #1}%
  \@@showhyphens{#1}%
}
\makeatother

\usepackage[utf8]{inputenc}
\usepackage[T1]{fontenc}
\usepackage{float}
\title{Real-Time Cell Sorting with Scalable In Situ FPGA-Accelerated Deep Learning}

\author[1]{Khayrul Islam}
\author[2]{Ryan F. Forelli}
\author[3]{Jianzhong Han}
\author[4]{Deven Bhadane}
\author[3,5,6]{Jian Huang}
\author[7]{Joshua C. Agar}
\author[8]{Nhan Tran}
\author[2]{Seda Ogrenci}
\author[9,*]{Yaling Liu}

\affil[1]{Department of Mechanical Engineering, Lehigh University, Bethlehem, PA 18015, USA}
\affil[2]{Department of Electrical and Computer Engineering, Northwestern University, Evanston, IL 60208, USA}
\affil[3]{Coriell Institute for Medical Research, Camden, NJ, USA}
\affil[4]{Department of Computer Science, Lehigh University, Bethlehem, PA 18015, USA}
\affil[5]{Cooper Medical School of Rowan University, Camden, NJ 08103, USA}
\affil[6]{Center for Metabolic Disease Research, Temple University Lewis Katz School of Medicine, Philadelphia, PA 19122, USA}
\affil[7]{Department of Mechanical Engineering and Mechanics, Drexel University, Philadelphia, PA 19104, USA}
\affil[8]{Real-time Processing Systems Division, Fermi National Accelerator Laboratory, Batavia, IL 60510, USA}
\affil[9]{Department of Bioengineering, Lehigh University, Bethlehem, PA 18015, USA}

\affil[*]{yal310@lehigh.edu}

\keywords{CNN, Cell Sorting, FPGA, Computer Vision, Knowledge Distillation}

\begin{abstract}
Precise cell classification is essential in biomedical diagnostics and therapeutic monitoring, particularly for identifying diverse cell types involved in various diseases. Traditional cell classification methods, such as flow cytometry, depend on molecular labeling, which is often costly, time-intensive, and can alter cell integrity. Real-time microfluidic sorters also impose a sub-ms decision window that existing machine-learning pipelines cannot meet. To overcome these limitations, we present a label-free machine learning framework for cell classification, designed for real-time sorting applications using bright-field microscopy images. This approach leverages a teacher--student model architecture enhanced by knowledge distillation, achieving high efficiency and scalability across different cell types. Demonstrated through a use case of classifying lymphocyte subsets, our framework accurately classifies T4, T8, and B cell types with a dataset of 80,000 pre-processed images, released publicly as the \texttt{LymphoMNIST} package for reproducible benchmarking. Our teacher model attained 98\% accuracy in differentiating T4 cells from B cells and 93\% accuracy in zero-shot classification between T8 and B cells. Remarkably, our student model operates with only 5682 parameters ($\sim$0.02\% of the teacher, a 5000-fold reduction), enabling field-programmable gate array (FPGA) deployment. Implemented directly on the frame-grabber FPGA as the first demonstration of in-situ deep learning in this setting, the student model achieves an ultra-low inference latency of just 14.5~$\mu$s and a complete cell detection-to-sorting trigger time of 24.7~$\mu$s, delivering 12$\times$ and 40$\times$ improvements over the previous state of the art in inference and total latency, respectively, while preserving accuracy comparable to the teacher model. This framework establishes the first sub-25~$\mu$s ML benchmark for label-free cytometry and provides an open, cost-effective blueprint for upgrading existing imaging sorters.
\end{abstract}

\begin{document}

\flushbottom
\maketitle
%
%
\thispagestyle{empty}

\section{Introduction}
\label{sec:intro}

Accurate cell classification is critical for a wide range of biomedical applications, including disease diagnostics, immunological studies, and personalized therapies. Traditional methods for cell classification, such as molecular labeling through flow cytometry, rely on detecting specific surface markers \cite{MacLaughlin2013-eq}. While these techniques are accurate, they have notable limitations, including high costs, time-intensive protocols, and potential interference with the natural state of the cells \cite{Nawaz2020-bh}. Equally important, modern acoustofluidic sorters provide only a $\sim$1 ms window between image acquisition and actuation, a latency budget that no published machine-learning (ML) pipeline has yet satisfied\cite{Nawaz2023-xx}. In response to these challenges, label-free classification methods have emerged as a promising alternative by leveraging intrinsic cell properties, such as morphology and biomechanics. Recent work has demonstrated the fundamental interconnection between biophysical cues and cellular morphology, with substrate geometry alone capable of reverting pluripotent stem cells to naivety through morphological changes\cite{Xu2024-cf}. This highlights that morphological features captured in bright-field images encode meaningful information about cellular state and phenotype. These approaches preserve the natural state of the cells, enabling downstream applications such as transplantation, functional studies, and real-time analysis or sorting \cite{Wang2020-pb, Islam2025-dp}.

Recent advancements in ML have revolutionized cell classification by offering innovative solutions to circumvent the limitations of traditional methods. For instance, deep CNNs have been successfully applied to bright-field images for label-free identification of cell types, with feature fusion approaches integrating morphological patterns across multiple convolutional modules to achieve high accuracy \cite{Iqbal2025-rz}. While such specialized approaches show promise, many general ML models perform suboptimally on datasets other than those they were specifically trained on, revealing inadequate generalization and transfer-learning capabilities. Furthermore, training protocols optimized for general image datasets often fail to translate effectively to biological datasets \cite{Zhang2020-qs,Sytwu2022-ct}. Progress is also slowed by the scarcity of large, publicly available bright-field datasets with ground-truth phenotypes, making reproducible benchmarking difficult.

\begin{figure}[htbp]
\centering
\includegraphics[width=.85\linewidth]{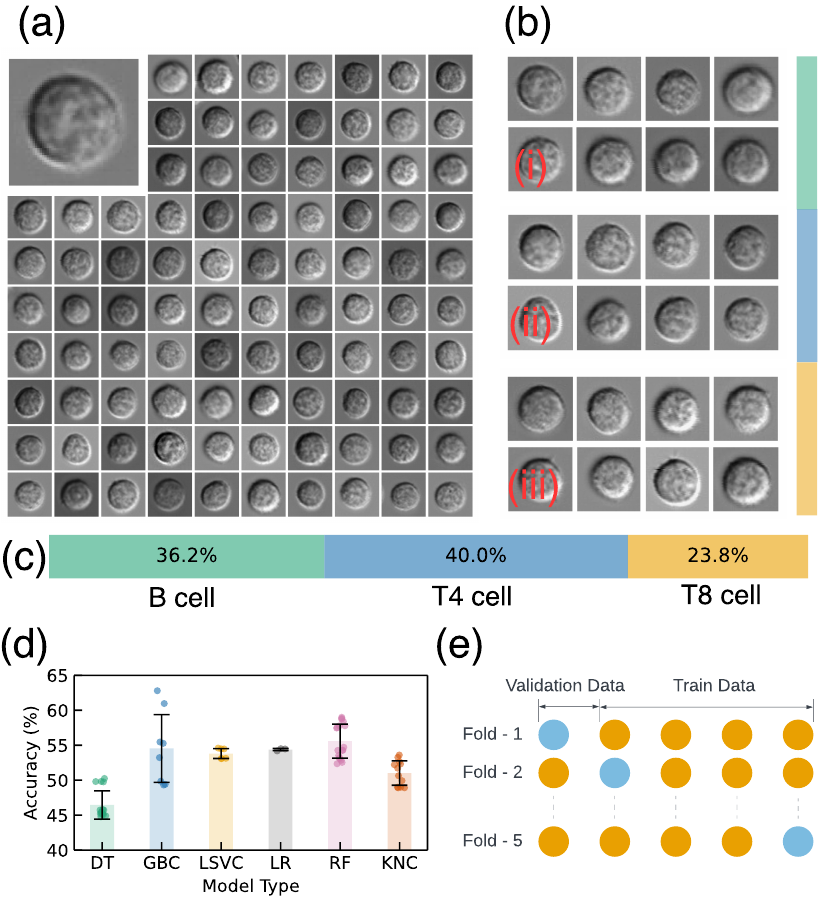}
\caption{Overview of LymphoMNIST dataset and ablation study. (a) sample dataset images, (b) examples of B(I), T4(II), and(III) T8 cells, (c) cell type distribution, (d) classification method performance in ablation study, and (e) training-validation split visualization.
}\label{fig1}
\end{figure}

Addressing these challenges requires robust training methodologies tailored specifically for diverse biological image datasets. In this study, we focus on optimized training protocols that achieve high specificity and sensitivity in cell classification. Using lymphocyte classification as a use case, we demonstrate the adaptability and effectiveness of these training recipes, highlighting their potential to extend seamlessly to various cell types and enabling versatile applications across different biological contexts. Specialized expertise in lymphocyte classification remains limited even in well-resourced communities, leading to variability in diagnostic accuracy. This issue is exacerbated in underserved areas, where the lack of access to expert pathology services results in prolonged or erroneous diagnostic outcomes that critically impair patient management. Our ML framework leverages bright-field images to detect cellular morphological features for the cell classification process. By eliminating reliance on molecular labels, this approach reduces human subjectivity, ensures reproducibility, and offers consistent results across different settings. To facilitate community adoption, we release both our training code and the 80\,000-image \textit{LymphoMNIST} dataset as pip-installable packages.

Moreover, to meet the demands of real-time inference, we have implemented a field-programmable gate array (FPGA) version of our optimized student model, achieving ultra-low latency and high throughput. Previous studies have demonstrated cell classification ML inference performance on the order of milliseconds, primarily on GPU and CPU hardware \cite{Tang_low_latency, Gu_cell_sort, Li_2019_DL}. The previous state-of-the-art (SOTA) in terms of inference latency implements a deep neural network (DNN) for standing surface acoustic wave cell sorting and achieves an inference latency of approximately 183\,\textmu s \cite{Nawaz2020-bh} and a full cell detection-to-sorting trigger latency of $<1$\,ms. Leveraging high-level-synthesis tools (\texttt{hls4ml}) and a knowledge-distilled student network with only 5682 parameters (about 0.02 \% of the 28 M-parameter teacher, a 5000-fold reduction), we achieve the first frame-grabber-resident deep-learning implementation that fits within this strict latency envelope.

By leveraging our ML framework in a use case involving the classification of T4, T8, and B cells, we have achieved remarkable accuracy improvements. Our teacher model demonstrates approximately 98 \% accuracy in classifying T4 cells from B cells and achieves about 93 \% accuracy in zero-shot classification of T8 vs.\ B cells. Employing knowledge-distillation (KD) techniques, our Student 2 model attains sufficiently high accuracy relative to the teacher model with just about 0.02 \% of its parameters. The FPGA implementation of the student model further enhances processing speed, reducing inference latency to just 14.5\,\textmu s. This improvement in processing speed facilitates the real-time analysis and accurate sorting of T and B cells, significantly advancing their rapid classification in clinical settings. With these insights and results in place, the core achievements and contributions of our study are summarized in the following research highlights:

\begin{enumerate}
    \item \textbf{\texttt{Dataset}}: We present a dataset of 80000 images, which supports the training and validation of our models. The data are freely available via the pip-installable \textit{LymphoMNIST} package for immediate benchmarking.

    \item \textbf{\texttt{Models}}: We publish detailed recipes for a high-capacity teacher and a KD-trained student with an in-house, lightweight architecture tuned for bright-field cell images, achieving 5000-fold parameter compression (5682 params, 0.02 \% of the teacher) while retaining F1\,$>$\,0.97.

    \item \textbf{\texttt{Transfer Learning}}: We demonstrate the transfer-learning capability for T8 versus B cell classifications, indicating that the model can perform zero-shot inference and can be further tuned to detect other lymphocyte cell types.

    \item \textbf{\texttt{In-Situ FPGA Implementation}}: We deploy our student model directly on the frame-grabber FPGA, eliminating PCIe transfer overhead and reducing inference latency from the 183\,\textmu s previous SOTA and 325\,\textmu s on GPU to just 14.5\,\textmu s, a 12× and 22× improvement, respectively. Thus, we institute a new SOTA real-time deep-learning benchmark and implementation for real-time cell sorting and rapid classification.
\end{enumerate}

\section{Results and Discussion}
\label{sec:Results}

\subsection{Composition of Training and Validation Sets}

The LymphoMNIST dataset consists of 80,000 high-resolution lymphocyte images, each with a resolution of 64×64 pixels (Figure \ref{fig1}(a)). These images are categorized into three primary classes: B cells, T4 cells, and T8 cells (Figure \ref{fig1}(b,c)). To support the development and evaluation of machine learning models, the dataset is partitioned into training, validation, and testing sets in an 80-10-10 split, resulting in 64,000 images for training and 8,000 images each for testing and validation (Figure \ref{fig1}(e)). To enhance accessibility and usability, we have developed a pip-installable package that allows researchers to seamlessly download the dataset and incorporate it into their experimental workflows \cite{LymphoMNIST}. The images in the dataset were captured under diverse environmental conditions, including variations in lighting and camera settings, to introduce a realistic level of complexity for algorithm development. These conditions are designed to simulate the variability encountered in real-world scenarios, challenging models to generalize effectively. Furthermore, the dataset includes images from both young (65\%) and aged (35\%) mice to account for age-specific cellular variability, a factor that enhances the model's ability to generalize across different biological conditions. The collection spanned 18 months across four seasons, ensuring that environmental fluctuations such as controlled humidity (\(\pm 5\%\)) and temperature (\(\pm 2^\circ\)C) were captured, further contributing to dataset diversity. Performance benchmarks of various models, like Decision Tree (DT), Gradient Boosting Classifier (GBC), Linear Support Vector Classifier (LSVC), Logistic Regression (LR), Random Forest (RF), and K-Nearest Neighbors Classifier (KNC), applied to the dataset are detailed in the supplementary materials. Accuracy metrics for these models are presented in Figure \ref{fig1}(d), providing insights into the dataset's applicability for machine learning tasks.

\subsection{Detection of Cell Class by Teacher:}

In this study, we utilized the ResNet50 architecture as our Teacher Network (TN) for the classification of B cells and T4 cells using bright-field microscopy images. ResNet50 is a deep convolutional neural network with residual connections, designed to alleviate the vanishing gradient problem and enable deeper feature extraction. Its capability to learn hierarchical representations makes it well-suited for complex image classification tasks such as distinguishing cell types \cite{He2015-gi}.

We observed that increasing the image size from the original \(64 \times 64\) pixels in the LymphoMNIST dataset to \(120 \times 120\) pixels improved both training and validation accuracy. This larger size allowed TN to capture more spatial information, enhancing feature extraction. The choice of image size is closely tied to the depth of the architecture, as deeper models like ResNet50 can leverage larger feature maps for improved performance, as noted in previous research \cite{Rahimzadeh2024-qj,Saponara2022-dz}. However, further increasing the size led to overfitting due to the model’s increased complexity. Thus, we standardized all images to \(120 \times 120\) pixels to achieve an optimal balance between feature learning and generalization.

\begin{figure}[htbp]
\centering
\includegraphics[width=.95\linewidth]{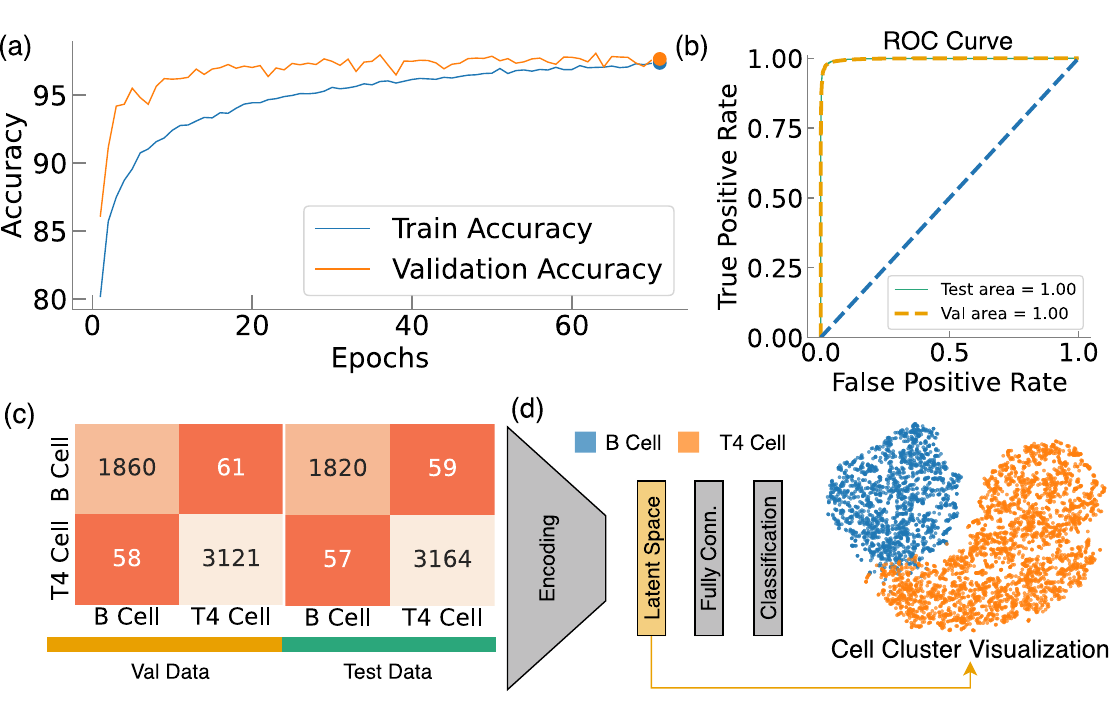}
\caption{Evaluation of the Teacher Network (TN). (a) accuracy during training and testing phases, (b) ROC curve, (c) confusion matrix demonstrating model efficacy on training and validation datasets using the TN, (d) depiction of the TN framework and t-Distributed Stochastic Neighbor Embedding (t-SNE) visualization derived from the latent space.
}\label{fig2}
\end{figure}

To improve generalization and reduce overfitting, we employed a range of data augmentation techniques, including random flips, rotations, scaling, translations, shearing, contrast adjustments, hue and saturation adjustments, and Gaussian blur. The choice and intensity of these augmentations must be carefully balanced depending on both the complexity of the model and the amount of available data. For complex models like ResNet50, stronger augmentations can introduce sufficient variability, preventing the model from overfitting by helping it generalize better across the dataset~\cite{Ethiraj2022-ps}. However, when the dataset is limited, applying overly strong augmentations can introduce excessive noise, which may degrade performance, particularly in tasks with high-dimensional latent spaces (Figure \ref{fig2}a) by causing the model to fit irrelevant or spurious patterns~\cite{Dablain2023-xn}. In such cases, it can be more effective to use a less complex model that is better suited to the smaller dataset, as it reduces the risk of overfitting to noise and irrelevant patterns in the training data~\cite{Faghri_undated-qq}. The dataset exhibited a class imbalance between B cells and T4 cells. To address this, we employed a weighted random sampler during training to ensure that the underrepresented classes were adequately sampled. This approach allowed the model to learn distinguishing features for both classes effectively, preventing bias towards the majority class (Figure \ref{fig2}c).

\begin{table}[htbp]
  \centering
  \caption{Comparison of Model Performance with Published Studies.}
  \small
  \resizebox{0.5\textwidth}{!}{%
    \begin{tabular}{@{}l l l l l@{}}
      \toprule
      Study & Imaging Technique & Model & Metric & Score \\
      \midrule
      Turan et al. \cite{Turan2018-pg} & Fluorescence & AlexCAN & Accuracy & 98\% \\
      Nassar et al. \cite{Nassar2019-ll} & Bright-field & Gradient Boosting & F1 Score & 78\% \\
      \midrule
      \textbf{This Study} & \textbf{Bright-field} & \textbf{Teacher} & \textbf{Accuracy} & \textbf{98\%} \\
      & & & \textbf{F1 Score} & \textbf{97.05\%} \\
      \bottomrule
    \end{tabular}%
  }
  \label{tab1}
\end{table}

The TN model achieved a training accuracy of approximately 97\%, and a validation accuracy of approximately 98\% after 70 epochs (Figure \ref{fig2}a). The close alignment between the training and validation accuracies indicates strong generalization without significant overfitting. Notably, the validation accuracy occasionally surpassed the training accuracy, likely due to the extensive augmentations applied to the training data, which were not applied to the validation set. Figure \ref{fig2}(b) shows the Receiver Operating Characteristic (ROC) curve, which highlights the model's strong discriminatory capability between B cells and T4 cells, with a high Area Under the Curve (AUC) for both the training and validation datasets. The confusion matrix in Figure \ref{fig2}(c) demonstrates high true positive rates and low false positive rates for both classes. Finally, the t-distributed Stochastic Neighbor Embedding (t-SNE) visualization (Figure \ref{fig2}d) provides a visual representation of the separation between B cells and T4 cells in the latent feature space. The minimal overlap between clusters further confirms the model's ability to effectively capture distinguishing features between the two cell types, making it a reliable tool for cell classification in biomedical applications.

\begin{figure}[htbp]
\centering
\includegraphics[width=.8\linewidth]{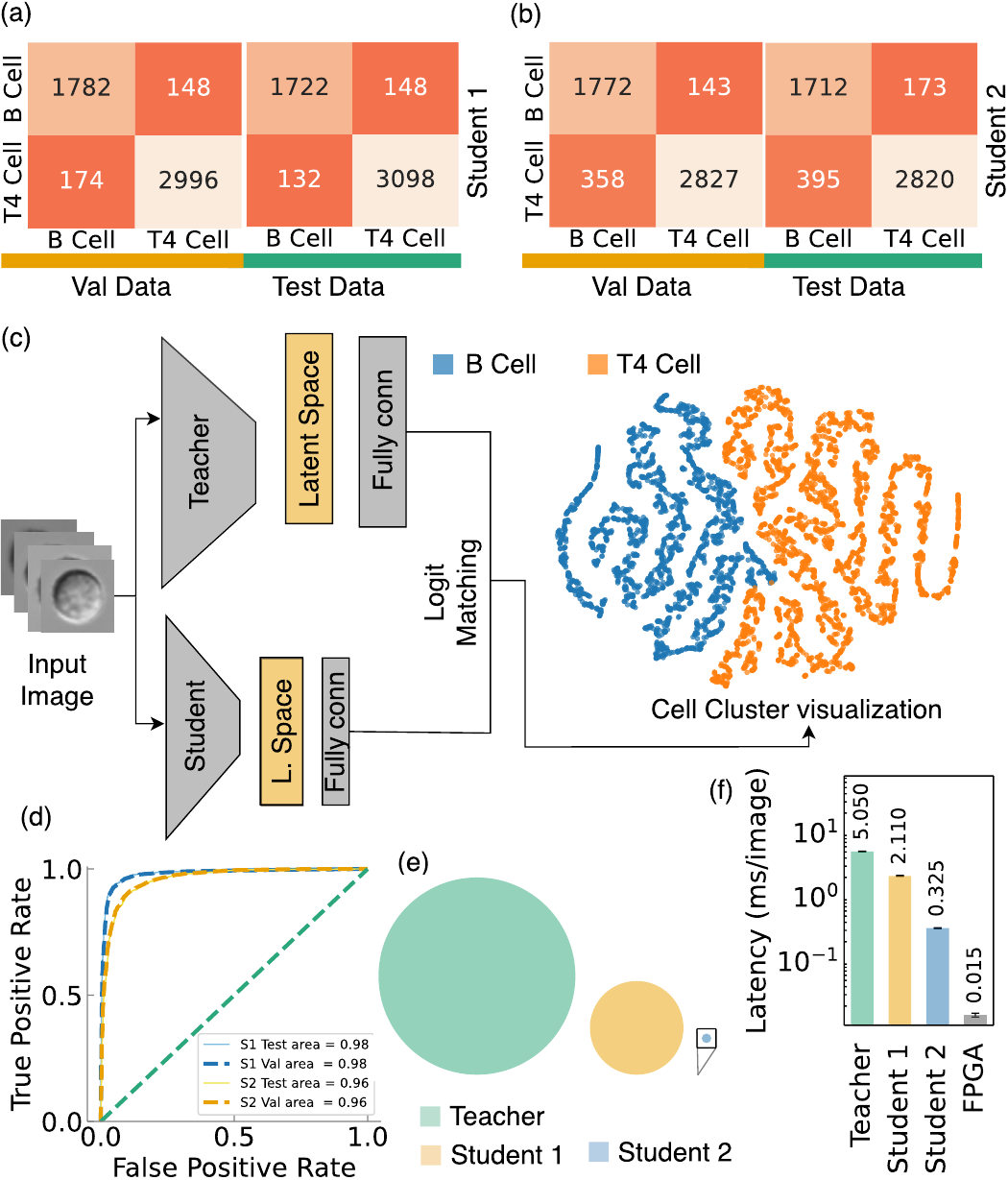}
\caption{Performance evaluation of the Student Network (SN). (a) confusion matrix for Student 1, (b) confusion matrix for Student 2, (c) t-SNE visualization of the SN framework, (d) ROC curve, (e) comparative analysis of model parameters (Student 2 magnified 200x), (f) latency comparison between teacher and student networks.}
\label{fig3}
\end{figure}

\subsection{Detection of Cell Class by Student:}

In this section, we investigate the effectiveness of KD in training student models by transferring knowledge from a pre-trained teacher model. Adopting the principles from Beyer et al. \cite{Beyer2021-px}, we employed a ``consistent and patient'' teaching strategy, emphasizing the importance of long training schedules and uniform input views between teacher and student. The distillation process allows the student model to leverage the richer representations of the teacher, improving its predictive capabilities. In this study, we trained two distinct student models, referred to as Student 1 and Student 2. Student 1 utilizes ResNet-18, a moderately complex convolutional neural network (CNN) with approximately 11.2 million parameters and an input size of \(64 \times 64\) pixels. We also developed a significantly compact model, Student 2, which is a lightweight CNN optimized for resource-constrained devices with only 5,682 parameters and a smaller input size of \(48 \times 48\). Notably, Student 2 achieved approximately 90\% accuracy in the classification task, demonstrating high efficiency with just 0.02\% of the parameters used by the teacher model, which achieved $\sim$98\% accuracy.

Our experiments reconfirmed that data-mixing augmentation techniques, such as CutMix and MixUp, substantially enhance KD performance. Conversely, other image-based augmentations, including random flipping and shearing, degraded the accuracy of the distilled student model when applied inconsistently between teacher and student \cite{Wang2020-zl}, as demonstrated by Beyer et al.~\cite{Beyer2021-px}. Maintaining identical image crops and augmentation strategies for both teacher and student networks during training was crucial to ensure consistent learning and effective knowledge transfer without misalignment in data representation \cite{Beyer2021-px}.

We observed that the Student 2 model attained significantly higher accuracy when trained using KD compared to training from scratch. This outcome aligns with prior research indicating that KD enables smaller models to focus on relevant information by utilizing outputs from a larger teacher model, including softened labels, as guidance \cite{Ballout2024-gy}. Such guidance allows the student model to capture complex patterns by receiving nuanced data representations, which may be challenging to learn independently, especially in resource-constrained scenarios \cite{Ba2014-ha}. Furthermore, studies have demonstrated that KD improves the ability of student models to capture high-level abstractions that are difficult to learn without teacher supervision \cite{Romero2014-cy}. For instance, Hinton et al. \cite{Hinton2015-ih} showed that soft targets enhance student model performance by conveying richer information about class relationships.

The performance evaluation of student networks, shown in Figure \ref{fig3}, reveals their accuracy on training and validation datasets. Confusion matrices on Figure \ref{fig3}(a) and \ref{fig3}(b) indicate that Student 1 slightly outperforms Student 2, although Student 2 demonstrates strong generalization capabilities in more challenging classes, suggesting that KD effectively maintains robustness in smaller models \cite{Gou2021-hq}. Figure \ref{fig3}(c) presents a t-SNE visualization for Student 1, showing distinct clusters that signify successful feature extraction and class differentiation. ROC curves (Figure \ref{fig3}(d)) for both models illustrate high discriminative performance, with AUC values of 98\% for Student 1 and 96\% for Student 2 respectively. Comparative analysis of model parameters and latency in Figure \ref{fig3}(e) and \ref{fig3}(f) reveals that Student 2 operates with only 0.02\% of the teacher model's parameters, achieving a latency of $\sim 0.325 \pm 0.004$ ms. This is significantly lower than Student 1 ($\sim 2.11 \pm 0.03$ ms) and the teacher model ($\sim 5.05 \pm 0.06$ ms), with the FPGA implementation further reducing latency to $\sim 0.0145 \pm 0.001$ ms.

\subsection{Transfer Learning for T4 and T8 Cell Classification:}

\begin{figure}[htbp]
\centering
\includegraphics[width=1\linewidth]{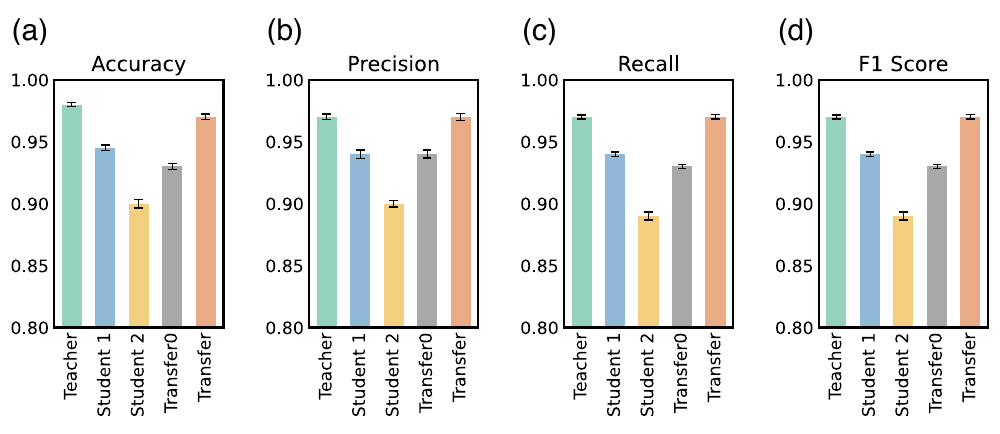}
\caption{Model Performance Evaluation. (a)-(d) present the comparative assessment of Accuracy, Precision, Recall, and F1 score.
}\label{fig4}
\end{figure}

This section investigates the utilization of transfer learning to differentiate between T8 and B cells employing a pre-trained teacher model. Originally trained on the LymphoMNIST dataset, the teacher network exhibited substantial feature extraction capabilities, achieving $\sim$98\% accuracy on both validation and test datasets. In a zero-shot learning framework (Transfer0 in Figure \ref{fig4}) for classifying T8 versus B cells, the model demonstrated an initial accuracy of $\sim$93\%. To improve classification performance, the teacher model underwent fine-tuning on a subset of the dataset specifically annotated for T8 and B cells. This fine-tuning involved modifying the training regimen to include only eight epochs, which facilitated model convergence without inducing overfitting. Post fine-tuning, the model reached an improved accuracy of $\sim$97\%, which surpassed its zero-shot learning performance.

To further assess the generalizability of the transfer learning approach beyond the specific T8 vs. B cell classification task, we evaluated our model on an external dataset \cite{Jin2023-be}, which includes additional hematological cell types. Our results demonstrated a \textasciitilde1\% accuracy boost for T vs. Leukemia cell classification when using our pretrained teacher model as the starting point, compared to an ImageNet-pretrained ResNet50. This indicates that leveraging prior domain-specific knowledge enhances model adaptability across different cell types and pathological conditions, reinforcing the robustness of our transfer learning strategy.

Figure \ref{fig4} illustrates the model's performance through comparative assessments of Accuracy, Precision, Recall, and F1 score across panels (a) to (d). The adaptability of the model to the new classification task, with minimal risk of overfitting and improved generalization capabilities, highlights the practical application of transfer learning in biomedical image analysis. Future research directions include extending these methodologies to other cell types or imaging modalities and combining them with continuous learning strategies or domain adaptation to enhance model performance under diverse imaging conditions.

\subsection{FPGA Implementation of the Student Model:}

For real-time cell sorting applications, latency is more critical than throughput because a decision must be made quickly within the short period that each cell spends passing under the camera's region of interest after detection and before passing through the acoustic sorting region. GPUs are specifically designed for high throughput processing as they have high-bandwidth memory and can handle massive data flow. However, they falter with latency-sensitive tasks as they are not optimized for single-threaded performance. In our testing, Student 2 achieves an average inference latency of 0.325 ms and can reach a throughput of 3.1 kfps with a batch size of 1 on our NVIDIA A100 GPU.

\begin{figure}[htbp]
\centering

\includegraphics[width=.40\linewidth]{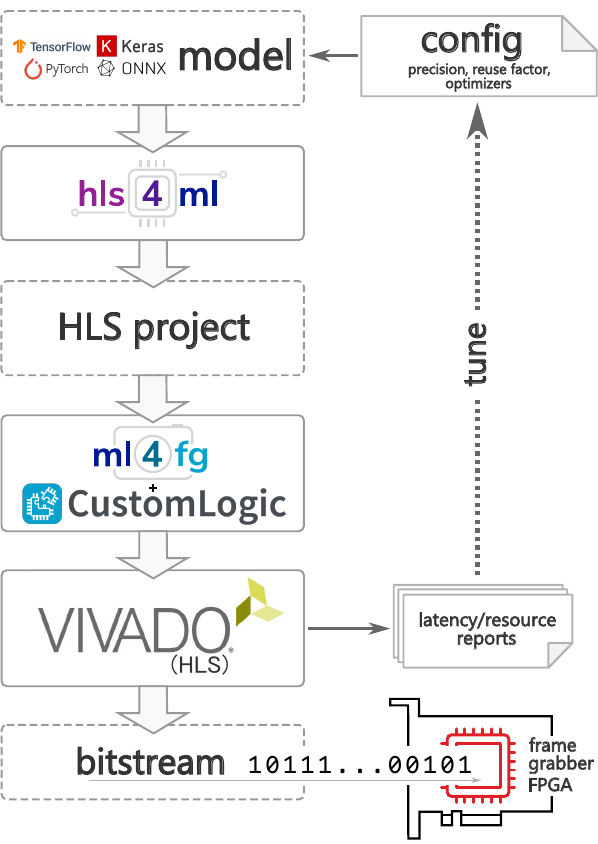}
\caption{Workflow from a high-level Python framework to HLS conversion, synthesis, frame grabber integration, and final bitstream generation of deep learning models with \texttt{hls4ml} for frame grabber deployment. The final generated bitstream contains the configuration information that the FPGA uses to implement our deep learning model.}
\label{hls4ml_flow}
\end{figure}

To achieve the latencies required for real-time control in cell sorting, an alternative platform is required. FPGAs are devices characterized by their flexibility and parallelism and provide a suitable balance between throughput and latency for real-time applications. They primarily consist of an array of reconfigurable hardware blocks, such as multipliers, logic blocks, and memories that can be used to implement an algorithm directly as a circuit, thereby forgoing the stack of software and drivers required for a GPU or CPU implementation. Additionally, the emergence of HLS technologies, enabling the synthesis of standard C++ code to register-transfer level hardware descriptions, means that deploying algorithms to custom hardware is easier than ever.

\begin{table}[ht]
\centering
\scriptsize
\setlength{\tabcolsep}{3pt} 
\begin{tabular}{@{}lcccc@{}}
\toprule
\textbf{Method}                & \textbf{Accuracy} & \textbf{Latency}       & \textbf{Platform}  & \textbf{App.}      \\ \midrule
Ours                           & 86\%              & 14.5$\mu$s             & FPGA               & Cell Sort.                       \\
Ours                           & 90\%              & 325$\mu$s              & GPU                & Cell Sort.                       \\
Prior SOTA \cite{Nawaz2020-bh} & -                 & 183$\mu$s              & CPU                & Cell Sort.                       \\
CellSighter \cite{Pang2024-cy} & 88–93\% (Recall)  & -                      & GPU                & Cell Class.                      \\
FPGA DL \cite{Mouri-Zadeh-Khaki2025-zo} & 89.5\%       & 652$\mu$s               & FPGA               & Obj. Class.                      \\ \bottomrule
\end{tabular}
\vspace{-.2cm}
\caption{Comparison of our method with other SOTAs.}
\label{tab:2}
\end{table}

Furthermore, tools like \texttt{hls4ml} facilitate the process of deploying neural networks to FPGA hardware and have been shown capable of achieving nanosecond-level latencies for machine learning inference \cite{hls4ml_subnano_paper}. \texttt{hls4ml} enables the translation of most neural network architectures written in a high-level deep learning framework such as PyTorch or Keras/Tensorflow to an HLS representation using dictionary configuration files and prewritten layer templates for all common HLS synthesis tools including Xilinx, Intel, and Siemens \cite{pt_paper,tf_paper,keras_sw}.

\texttt{hls4ml} provides multiple avenues of optimization that empowers us to meet this project's latency constraints. First and foremost, previous work has demonstrated that neural network parameters can be quantized to a lower bit width with minimal impact on overall accuracy~\cite{Hashemi2017-of}. This finding is critical for enabling the deployment of neural networks on resource-constrained devices. In this implementation, we use \texttt{hls4ml} to quantize the Student 2 network with layer-level granularity while still achieving 86\% accuracy. We also leverage \texttt{hls4ml}'s "reuse factor" hyperparameter to fine-tune the level of parallelization applied to each layer of the network. The value of this parameter indicates the maximum number of operations that can share a given physical instance of a resource. This feature allows us to achieve the ultra-low latencies required for this application while remaining within the resource constraints of the FPGA device. The effects of varying this hyperparameter can be illustrated as a Pareto frontier where a high reuse factor results in low resource usage but high latency, and a low reuse factor results in high resource usage but low latency~\cite{Wei2024_fusion}. In general, we find that implementing dense layers with a higher reuse factor of 25, and the two convolutional layers with lower reuse factors of 1 and 2, respectively, yields an optimal balance between latency and resource usage.

Apart from latency, another challenge to enabling real-time control presents itself in the substantial input/output (IO) overhead that we would incur when utilizing a CPU or any external PCIe GPU or FPGA accelerator. Therefore, we endeavored to place our Student 2 model computation as close to the edge as possible in our experiment to minimize this overhead. Our experimental setup consists of a Phantom S710 high-speed streaming camera aimed at the microfluidic channel through the microscope camera port, paired with the Euresys frame grabber PCIe card. This frame grabber card is responsible for reading out and processing the raw camera sensor data before transmitting frames back to the host computer. Frame grabbers typically implement this processing on an onboard FPGA chip. Conveniently, Euresys offers a tool, CustomLogic, that enables users to deploy custom image processing to their frame grabber FPGA~\cite{customlogic_software}. A separate framework, Machine Learning for Frame Grabbers (\texttt{ml4fg}) has also been developed specifically to bridge the gap between CustomLogic and \texttt{hls4ml} and enables seamless deployment of neural network models to Euresys frame grabbers~\cite{hls4ml_FG_RF}. Thus, we leverage all three of these existing tools to deploy Student 2 directly in situ in the data readout path of the frame grabber, thereby circumnavigating the need for off-chip compute and completely eliminating all associated IO overhead while achieving ultra-low latency inference. Our full workflow from Python model to bitstream deployment is illustrated in Figure \ref{hls4ml_flow}.

\begin{figure}[htbp]
\centering
\includegraphics[width=.85\linewidth]{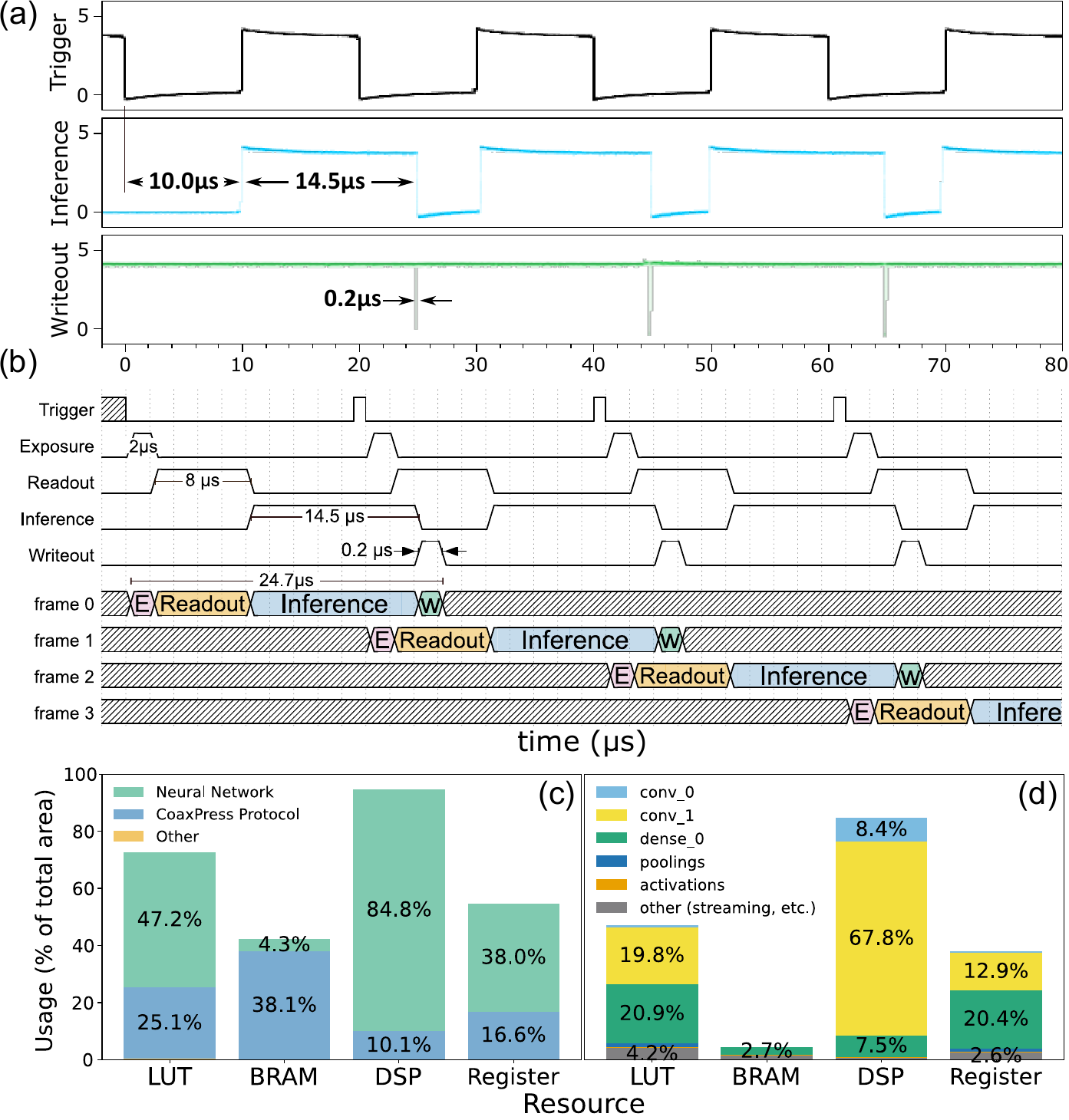}
\caption{Latency and resource performance of our FPGA implementation of Student 2. (a) Student 2 empirical oscilloscope benchmark of inference latency where “E” denotes camera exposure and “W” denotes the serial writeout, (b) Student 2 frame grabber inference timing diagram illustrating pipelined model inference with exposure and readout, (c) overall resource consumption of the FPGA broken down by IP, (d) resource consumption of the neural network IP broken down by layer.}
\label{fpga_results}
\end{figure}

We then empirically benchmark the latency of the FPGA implementation of Student 2 by monitoring the internal communication protocol used by the neural network intellectual property (IP). We then utilize the frame grabber's TTL IO to output a square wave where the high time denotes inference latency which we measure with an oscilloscope. Figure \ref{fpga_results}a exhibits the results of this latency test, showing a model inference latency of just 14.5~\mbox{$\mu$s}. Additionally, we observe that inference begins approximately 10.0~\mbox{$\mu$s} after the trigger edge. Given a 2~\mbox{$\mu$s} exposure time, our model completes inferencing approximately 22.5~\mbox{$\mu$s} after image exposure is finished. The model output writeout procedure takes an additional 0.2~\mbox{$\mu$s}. The writeout consists of the model's two-bit output indicating the cell output class, and can be expanded or adapted for any cell classification task or communication protocol. Aggregating these constituent components yields a full cell detection-to-sorting trigger time of 24.7~\mbox{$\mu$s}. By reducing inference latency to under 25~$\mu$s, our pipeline shifts the limiting factor from computation to fluidics. This margin not only exceeds the $\sim$1~ms actuation window of current acoustofluidic sorters \cite{Nawaz2023-xx}, but also opens the door to applications previously inaccessible to image-based ML—such as sorting extracellular vesicles or bacteria, where transit times are an order of magnitude shorter than for mammalian cells. As shown in Figure \ref{fpga_results}b, we pipeline neural network inference with the exposure and readout processes to accelerate the algorithm to a throughput of 81 kfps in our implementation. This benchmark far exceeds our GPU's best performance at a batch size of 1. Note that in Figure \ref{fpga_results}a we capture at 50 kfps such that consecutive inference traces do not overlap for readability purposes.

As shown in Figure \ref{fpga_results}c, our implementation of Student 2 consumes the majority of the FPGA resources. DSPs, the resource primarily used to implement neural network multiply accumulate operations, are most heavily utilized because we parallelized the network to the limit of the chip's resource capacity with \texttt{hls4ml}'s reuse factor hyperparameter. The high-speed camera's communication protocol IP imposes an additional resource tax, totaling about 95\% DSP usage for the full design. A more granular breakdown of the neural network resource consumption is shown in Figure \ref{fpga_results}d. Most notably, the second convolutional layer consumes far more resources than any other layer due to the higher number of input channels, which results in more multiply-accumulates. Both convolutional layers consume the most lookup tables as they require more complex control logic to manage the sliding kernel window and to direct data between buffers.

By optimizing our Student 2 model and leveraging existing tools like \texttt{hls4ml} for deployment in situ on low-cost off-the-shelf frame grabber FPGAs, we are able to bypass data transfer overhead and accelerate our deep learning algorithm to achieve a new SOTA 14.5~\mbox{$\mu$s} inference latency and 24.7~\mbox{$\mu$s} full cell detection-to-sorting trigger time for cell classification in real-time sorting applications(see Table~\ref{tab:2}).

\section{Methods}

\subsection{Animals}
Evi1-IRES-GFP knock-in (Evi1$^{\text{GFP}}$) mice, kindly provided by Dr. Mineo Kurokawa at the University of Tokyo, were used for this study.  The mice were bred and housed under specific-pathogen-free (SPF) conditions within the animal facility at Cooper University Health Care. All animal handling and experimental protocols adhered strictly to NIH-mandated guidelines for laboratory animal welfare. Protocols were reviewed and approved by the Institutional Animal Care and Use Committee (IACUC) at Cooper University Health Care to ensure compliance with ethical standards for the care and use of laboratory animals.

\begin{figure}[htbp]
\centering
\includegraphics[width=.9\linewidth]{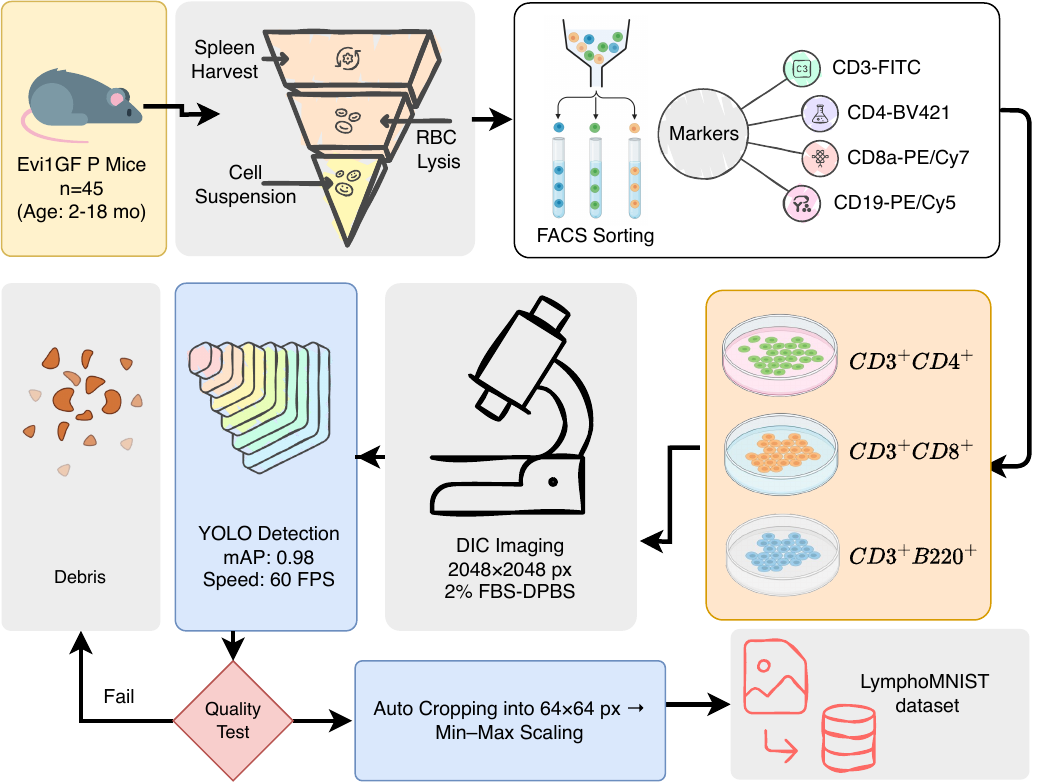}
\caption{Flowchart of LymphoMNIST data collection and preprocessing. Splenocytes were FACS-sorted into T4, T8, and B subsets, imaged by DIC microscopy, cropped with YOLOv5, pathologist-verified, and standardized to 64$\times$64 pixels for dataset generation.}
\label{fastml_flow}
\end{figure}

\subsection{Antibodies}
For flow cytometric analysis and cell sorting, the following fluorochrome-conjugated antibodies were used: CD3$\varepsilon$-FITC (BioLegend, cat\# 152304), CD4-BV421 (BioLegend, cat\# 100543), CD8a-PE/Cy7 (BioLegend, cat\# 100722), CD19-PE/Cy5 (eBioscience, cat\# 15-0193-82), and B220 (eBioscience, cat\# 56-0452-82). These antibodies were selected for their specificity in targeting key immune cell surface markers, enabling accurate discrimination of immune cell subpopulations through fluorescence-based gating strategies.

\subsection{Flow Cytometric Analysis and Cell Sorting}
Murine lymphocytes were isolated from spleen tissue. Spleens were carefully dissected and homogenized to produce single-cell suspensions, followed by red blood cell lysis to ensure a clear lymphocyte population. After washing with Dulbecco's Phosphate-Buffered Saline (DPBS), cells were stained with the selected fluorochrome-conjugated antibodies at 4$^\circ$C for 15--30 minutes to ensure optimal labeling conditions. Flow cytometric analysis and fluorescence-activated cell sorting (FACS) were performed using a Sony SH800Z automated cell sorter or a BD FACSAria\texttrademark{} III cell sorter. Negative controls were prepared with unstained cells to set appropriate gating thresholds. Data analysis was conducted using FlowJo software (v10) or the native software associated with the Sony cell sorter, employing stringent gating strategies to accurately identify and isolate specific immune cell subsets while excluding debris and non-viable cells.

\subsection{DIC Image Acquisition}
Following FACS, sorted cells were seeded into coverglass-bottomed chambers (Cellvis) and maintained in DPBS supplemented with 2\% fetal bovine serum (FBS) to preserve cell viability throughout the imaging process. Differential interference contrast (DIC) imaging was performed on an Olympus FV3000 confocal microscope, with images captured at a resolution of 2048$\times$2048 pixels. High-resolution DIC images allowed for precise morphological characterization of the cells. Additionally, simultaneous fluorescence imaging was conducted to verify the accuracy of the cell sorting. Consistent imaging conditions were maintained across sessions to facilitate comparability of the acquired images.

\subsection{Data Processing}

In this study, automated cell detection and image processing were conducted using the YOLOv5 object detection framework. Given the challenges posed by bright-field microscopy images, such as overlapping cells, debris, and lighting artifacts, YOLOv5 demonstrated exceptional accuracy and efficiency, achieving 98\% detection accuracy on our validation subset compared to 82\% for Watershed-based segmentation. This ensured a reliable dataset with minimal preprocessing errors that could impact downstream classification. YOLOv5 also automated the cropping process, reducing manual labor by over 300 hours, whereas traditional methods like thresholding and Watershed segmentation required manual correction for 30\% of images in pilot tests, introducing variability and delays. YOLOv5 efficiently identified and cropped individual cells from DIC images, standardizing each to 64×64 pixels centered on the cell, minimizing variability for downstream machine learning tasks. Its feature extraction capabilities detected cells despite variations in size, shape, or orientation, enabling high-throughput processing. Manual inspection filtered misidentifications like debris or clusters, ensuring only correctly identified T4, T8, and B cells were retained. This workflow balanced efficiency and accuracy, yielding 80,000 images split into training, testing, and validation sets as described in Results.

\section{Conclusion}

We have developed a label-free machine learning framework for the classification of lymphocytes—specifically T4, T8, and B cells—using bright-field microscopy images. Utilizing a teacher-student model architecture with knowledge distillation, we achieved high accuracy while significantly reducing model complexity. In future work, we will extend the model by implementing additional FPGA hardware for the object detection component, as the current version only focuses on object classification. This hardware integration will enable real-time, high-throughput lymphocyte detection and sorting, enhancing its utility in clinical settings. Furthermore, expanding the model to classify rare lymphocyte subsets or those involved in specific diseases may increase its clinical relevance. This framework presents a significant advancement and new SOTA in lymphocyte classification and general cell-sorting by offering a non-invasive, efficient, ultra-low latency, scalable solution. It provides a strong foundation for the development of automated, label-free cell sorting technologies for both research and clinical applications.

\section{Data Availability}
The LymphoMNIST dataset described in this work is publicly available to promote reproducibility and facilitate further research~\cite{lymphomnist_zenodo}. Researchers can readily access and integrate the dataset into their workflows using the Python package \texttt{LymphoMNIST}. Comprehensive documentation, installation guidelines, dataset exploration methods, and implementation examples are provided in the project's GitHub repository (\url{https://github.com/Khayrulbuet13/LymphoMNIST}).

\section{Code Availability}
All models and experiments described in this work were implemented using Python with PyTorch as the core deep learning framework. Complete scripts, model training recipes, and instructions necessary to reproduce the experiments and results reported in this study are openly accessible~\cite{lymphoml_zenodo} via \url{https://github.com/Khayrulbuet13/LymphoML}.

\section{Acknowledgements}
This work was supported by the National Science Foundation (NSF) grant number 2215789.

\bibliography{main}




\renewcommand{\thetable}{S\arabic{table}} 
\renewcommand{\thefigure}{S\arabic{figure}} 
\renewcommand{\thesection}{S\arabic{section}} 
\setcounter{section}{0}
\setcounter{table}{0}
\setcounter{figure}{0}


\clearpage
\setcounter{page}{1}


\baselineskip=2.5\baselineskip  

\begin{center}
    {\Huge \textbf{Real-Time Cell Sorting with Scalable In Situ FPGA-Accelerated Deep Learning}} \\
    \vspace{.5cm} 
    {\huge {Supplementary Material}}
\end{center}

\vspace{1cm}


\baselineskip=1.2em  

\section{Benchmark Results}
\label{sec:benchmark_results}

Table \ref{tab:test_accuracy} shows a comprehensive comparison of classification performance on the LymphoMNIST and MNIST datasets, including the test accuracy achieved by common machine learning classifiers across a range of hyperparameter configurations. Notably, there is a large performance disparity between the two datasets which reflects the increased complexity of LymphoMNIST. For instance, the tree-based methods such as DecisionTreeClassifier performed significantly worse on LymphoMNIST compared to MNIST across the entire hyperparameter search space. GradientBoostingClassifier shows significant improvement in accuracy on MNIST as the model complexity (e.g., number of estimators) increases, while achieving modest gains on LymphoMNIST relative to DecisionTreeClassifier.

Furthermore, KNeighborsClassifier and RandomForestClassifier also achieve superior accuracy on MNIST, while their performance on LymphoMNIST demonstrates their difficulty in capturing complex features. LogisticRegression and LinearSVC also achieve only moderate accuracy on LymphoMNIST, further highlighting the necessity for specialized approaches. This comparative analysis shows the value of LymphoMNIST as a challenging benchmark for advancing ML models and provides a complementary perspective to MNIST’s utility in assessing general-purpose algorithmic performance. This exploration provides a foundation for future investigation into improving ML efficacy on datasets like LymphoMNIST.

\begin{table}[!htb]
\centering
\footnotesize
\caption{Comparison of Classification Accuracy on LymphoMNIST and MNIST Datasets.}\label{tab:test_accuracy}
\begin{tabular}{@{}p{5cm}p{7.5cm}rr@{}}

    \toprule
    \textbf{Classifier} & \textbf{Parameters} & \multicolumn{2}{c}{\textbf{Test Accuracy}} \\ 
    \cmidrule(lr){3-4}
    & & \textbf{Our Dataset} & \textbf{MNIST} \\
    \midrule
    \textbf{DecisionTreeClassifier} &  \scriptsize{criterion=gini, max\_depth=10, splitter=best}  & 0.498 & 0.866 \\
    & \scriptsize{criterion=gini, max\_depth=50, splitter=best}         & 0.453 & 0.877 \\
    & \scriptsize{criterion=gini, max\_depth=100, splitter=best}        & 0.452 & 0.879 \\
    & \scriptsize{criterion=gini, max\_depth=10, splitter=random}       & 0.498 & 0.853 \\
    & \scriptsize{criterion=gini, max\_depth=50, splitter=random}       & 0.449 & 0.873 \\
    & \scriptsize{criterion=gini, max\_depth=100, splitter=random}      & 0.457 & 0.875 \\
    & \scriptsize{criterion=entropy, max\_depth=10, splitter=best}      & 0.498 & 0.873 \\
    & \scriptsize{criterion=entropy, max\_depth=50, splitter=best}      & 0.454 & 0.886 \\
    & \scriptsize{criterion=entropy, max\_depth=100, splitter=best}     & 0.457 & 0.886 \\
    & \scriptsize{criterion=entropy, max\_depth=10, splitter=random}    & 0.502 & 0.861 \\
    & \scriptsize{criterion=entropy, max\_depth=50, splitter=random}    & 0.452 & 0.883 \\
    & \scriptsize{criterion=entropy, max\_depth=100, splitter=random}   & 0.451 & 0.881 \\
    \midrule
    \addlinespace
    \textbf{GradientBoostingClassifier} & \scriptsize{max\_depth=3, n\_estimators=10, loss=deviance}   & 0.502 & 0.846 \\
    & \scriptsize{max\_depth=10, n\_estimators=10, loss=deviance}  & 0.555 & 0.933 \\
    & \scriptsize{max\_depth=50, n\_estimators=10, loss=deviance}  & 0.498 & 0.888 \\
    & \scriptsize{max\_depth=3, n\_estimators=50, loss=deviance}   & 0.532 & 0.926 \\
    & \scriptsize{max\_depth=10, n\_estimators=50, loss=deviance}  & 0.610 & 0.964 \\
    & \scriptsize{max\_depth=3, n\_estimators=100, loss=deviance}  & 0.553 & 0.949 \\
    & \scriptsize{max\_depth=10, n\_estimators=100, loss=deviance} & 0.628 & 0.969 \\
    \midrule
    \addlinespace
    \textbf{KNeighborsClassifier} & \scriptsize{weights=uniform, n\_neighbors=1, p=1} & 0.489 & 0.955 \\
    & \scriptsize{weights=uniform, n\_neighbors=1, p=2} & 0.490 & 0.943 \\
    & \scriptsize{weights=distance, n\_neighbors=1, p=1} & 0.489 & 0.955 \\
    & \scriptsize{weights=distance, n\_neighbors=1, p=2} & 0.490 & 0.943 \\
    & \scriptsize{weights=uniform, n\_neighbors=5, p=1} & 0.506 & 0.957 \\
    & \scriptsize{weights=uniform, n\_neighbors=5, p=2} & 0.499 & 0.944 \\
    & \scriptsize{weights=distance, n\_neighbors=5, p=1} & 0.527 & 0.959 \\
    & \scriptsize{weights=distance, n\_neighbors=5, p=2} & 0.517 & 0.945 \\
    & \scriptsize{weights=uniform, n\_neighbors=9, p=1} & 0.523 & 0.955 \\
    & \scriptsize{weights=uniform, n\_neighbors=9, p=2} & 0.522 & 0.943 \\
    & \scriptsize{weights=distance, n\_neighbors=9, p=1} & 0.536 & 0.955 \\
    & \scriptsize{weights=distance, n\_neighbors=9, p=2} & 0.533 & 0.944 \\
    \midrule
    \addlinespace
    \textbf{LinearSVC} & \scriptsize{loss=squared\_hinge, C=1.0, penalty=l2, multi\_class=ovr}    & 0.531 & 0.912 \\
    & \scriptsize{loss=squared\_hinge, C=10.0, penalty=l2, multi\_class=ovr}   & 0.531 & 0.885 \\
    & \scriptsize{loss=squared\_hinge, C=100.0, penalty=l2, multi\_class=ovr}  & 0.531 & 0.873 \\
    \midrule
    \addlinespace
    \textbf{LogisticRegression} & \scriptsize{C=1.0, penalty=l2, multi\_class=ovr}          & 0.545 & 0.917 \\
    & \scriptsize{C=10.0, penalty=l2, multi\_class=ovr}         & 0.542 & 0.916 \\
    \midrule
    \addlinespace
    \textbf{RandomForestClassifier} & \scriptsize{criterion=gini, max\_depth=10, n\_estimators=10}       & 0.535 & 0.930 \\
    & \scriptsize{criterion=gini, max\_depth=50, n\_estimators=10}       & 0.525 & 0.948 \\
    & \scriptsize{criterion=gini, max\_depth=100, n\_estimators=10}      & 0.523 & 0.948 \\
    & \scriptsize{criterion=entropy, max\_depth=10, n\_estimators=10}    & 0.532 & 0.933 \\
    & \scriptsize{criterion=entropy, max\_depth=50, n\_estimators=10}    & 0.528 & 0.949 \\
    & \scriptsize{criterion=entropy, max\_depth=100, n\_estimators=10}   & 0.531 & 0.949 \\
    & \scriptsize{criterion=gini, max\_depth=10, n\_estimators=50}       & 0.543 & 0.945 \\
    & \scriptsize{criterion=gini, max\_depth=50, n\_estimators=50}       & 0.575 & 0.968 \\
    & \scriptsize{criterion=gini, max\_depth=100, n\_estimators=50}      & 0.578 & 0.967 \\
    & \scriptsize{criterion=entropy, max\_depth=10, n\_estimators=50}    & 0.543 & 0.947 \\
    & \scriptsize{criterion=entropy, max\_depth=50, n\_estimators=50}    & 0.578 & 0.967 \\
    & \scriptsize{criterion=entropy, max\_depth=100, n\_estimators=50}   & 0.575 & 0.968 \\
    \bottomrule
\end{tabular}
\end{table}

\section{FPGA Resource-Latency Trade-offs and Scalability Limits}
\label{sec:fpga_pareto}

\begin{figure}[ht]
\centering
\includegraphics[width=0.7\linewidth]{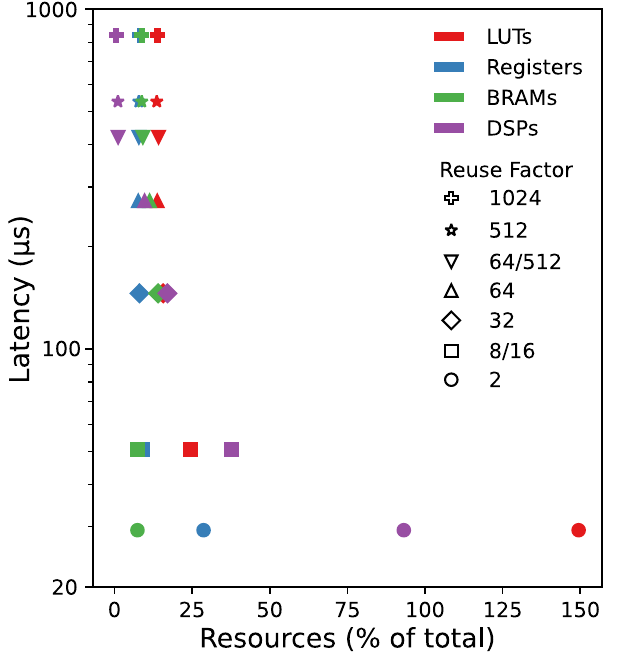}
\caption{Pareto frontier of optimized model's resource-latency configurations with increasing levels of parallelization. Colors represent the types of resources available on an FPGA: LUTs, Registers, BRAMs, and DSPs. Shapes represent different reuse factors. Resource usages are shown as \% of available resources on the KU035 FPGA. Adapted from Wei et al.\cite{Wei2024_fusion}.}
\label{fig:pareto_frontier}
\end{figure}

The reuse factor hyperparameter in \texttt{hls4ml} controls the degree of resource sharing in the synthesized hardware, directly affecting both latency and resource consumption. This trade-off can be visualized as a Pareto frontier, where design points represent different balance points between parallelization (low reuse factor, low latency, high resources) and resource efficiency (high reuse factor, high latency, low resources). Figure S\ref{fig:pareto_frontier} illustrates this trade-off for neural network FPGA implementations of comparable complexity to our Student~2 model. As reuse factor increases from 2 to 1024, latency increases from tens of microseconds to milliseconds while resource consumption (measured as percentage of available LUTs, registers, BRAMs, and DSPs on the KU035 FPGA) decreases dramatically. At reuse factor = 2, DSP utilization approaches 100\%, representing the low-latency extreme of the feasible design space. Our Student~2 implementation operates at this frontier, using reuse factors of 1--2 for convolutional layers and 25 for dense layers, achieving 14.5~\textmu s inference latency while consuming approximately 95\% of available DSPs.

Several factors constrain scalability to larger models or multi-class extensions on the target KU035 device. First, DSP exhaustion is the primary bottleneck: our design already consumes $\sim$95\% of available DSP blocks at near-minimum reuse factors. Extending to deeper networks or higher-resolution inputs (e.g., 64$\times$64 instead of 48$\times$48) would require either increasing reuse factors—thereby trading latency for feasibility—or adopting LUT-based arithmetic, which increases LUT consumption and may reduce maximum clock frequency. Second, memory bandwidth and capacity impose practical limits: aggressive parallelization demands more concurrent memory ports and greater BRAM partitioning, which can exceed on-chip resources for large models. Third, routing congestion becomes significant in highly parallelized designs, potentially degrading \(F_{\text{max}}\) such that even reduced cycle counts do not proportionally improve end-to-end latency. Finally, multi-class extensions increase computational and memory demands: for example, a 10-class classifier would require approximately 5$\times$ more DSPs and parameters in the final layer (a consequence of adding 8 more neurons) compared to our binary design. Our Student~2 architecture and heterogeneous reuse factor strategy represent a carefully optimized design point that achieves ultra-low latency (14.5~\textmu s inference, 24.7~\textmu s total detection-to-trigger) while remaining within the strict resource envelope of the KU035 FPGA. This positions our implementation at the practical performance limit for this device class and demonstrates the necessity of co-designing model architecture and hardware configuration to meet real-time sorting constraints.

\section{Model Calibration Analysis}
\label{sec:calibration_analysis}

Probability calibration measures whether a model's predicted confidence scores accurately reflect its true classification accuracy. A well-calibrated model assigns high probabilities to samples it classifies correctly and low probabilities to potential misclassifications. We evaluated calibration performance using Expected Calibration Error (ECE) and Maximum Calibration Error (MCE) with 5-bin reliability analysis~\cite{Guo2017-rk}. ECE measures the average difference between predicted confidence and actual accuracy across bins, while MCE captures the worst-case deviation. Temperature scaling~\cite{Guo2017-rk} was applied as a post-hoc calibration technique to assess whether simple rescaling of the output logits could improve calibration metrics.

\begin{figure}[ht]
\centering
\includegraphics[width=0.65\linewidth]{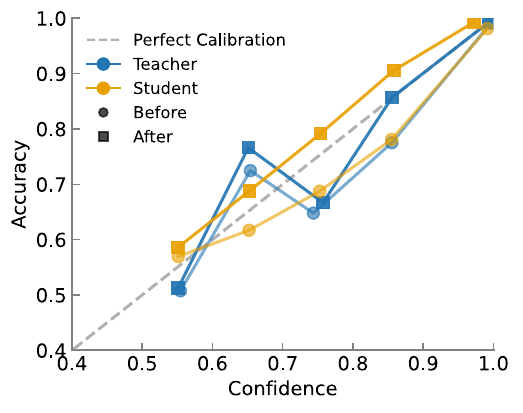}
\caption{Reliability diagram comparing Teacher (blue) and Student 1 (orange) model calibration before (circles) and after (squares) temperature scaling. Gray dashed line indicates perfect calibration.}
\label{fig:calibration_reliability}

\end{figure}

Figure~\ref{fig:calibration_reliability} presents reliability diagrams for the Teacher and Student~1 models before and after temperature scaling. The Teacher model demonstrates excellent baseline calibration (ECE = 0.90\%, MCE = 9.58\%), with post-hoc temperature scaling (optimal temperature $T = 1.286$) further improving ECE to 0.41\%—a 54.7\% reduction. The Student~1 model exhibits moderate baseline calibration (ECE = 1.59\%, MCE = 7.42\%) and displays underconfidence across low-to-mid confidence bins, meaning it tends to assign lower probabilities than its actual accuracy warrants. This conservative behavior is advantageous for classification tasks where false positives carry higher costs than false negatives, as underconfident predictions naturally provide a safety margin. Temperature scaling reduces worst-case calibration error (MCE from 7.42\% to 4.75\%), but increases average calibration error (ECE to 2.51\%), indicating that attempts to shift predictions toward perfect calibration disrupt the model's inherent conservative bias.

\section{Uncertainty-Aware Rejection and Throughput-Safety Trade-offs}
\label{sec:uncertainty_rejection}

Real-world deployment of machine learning models for cell sorting requires careful consideration of performance under domain shift and the ability to reject ambiguous predictions. In clinical settings, false routing of cells to incorrect collection channels can compromise downstream analyses or therapies. To address this critical deployment concern, we implemented and evaluated an uncertainty-aware rejection framework that quantifies the trade-off between throughput (fraction of samples processed) and safety (accuracy on processed samples).

Our rejection mechanism is based on confidence thresholding, where samples with maximum softmax probability below a threshold $\tau$ are rejected and not processed (directed to a discard channel in a physical sorting implementation). We evaluated seven threshold levels: $\tau \in \{0.50, 0.70, 0.80, 0.85, 0.90, 0.95, 0.99\}$ on both our Teacher and Student1 models. To simulate domain shift conditions that would be encountered when deploying the system across different microscopes, illumination conditions, or sample preparation protocols, we applied realistic augmentation transforms to our test dataset. These transforms included color jitter with brightness and contrast variations of 0.4 (simulating illumination changes), random rotations up to 30 degrees (simulating sample orientation variations), random horizontal flips with 50\% probability (simulating microscope positioning differences), and Gaussian blur (simulating optical variations). The test dataset consisted of 7,140 samples from the LymphoMNIST dataset, using a 70\% test split with random state 42 for reproducibility.

\begin{table}[ht]
\centering
\small
\caption{Uncertainty-aware rejection performance across confidence thresholds under clean and domain-shifted conditions.}
\label{tab:uncertainty_results}
\begin{tabular}{@{}lccccc@{}}
\toprule
\textbf{Model} & \textbf{Threshold} & \textbf{Coverage} & \textbf{Clean Acc.} & \textbf{Shift Acc.} & \textbf{False-Route Rate} \\
 & $\tau$ & (\%) & (\%) & (\%) & (Clean / Shift, \%) \\
\midrule
Teacher & 0.50 & 100.0 & 97.4 & 94.6 & 2.6 / 5.4 \\
 & 0.85 & 94.6 & 98.9 & 97.9 & 1.1 / 2.1 \\
 & 0.95 & 89.4 & 99.5 & 99.0 & 0.5 / 1.0 \\
\midrule
Student1 & 0.50 & 100.0 & 94.5 & 84.4 & 5.5 / 15.6 \\
 & 0.85 & 90.2 & 97.6 & 94.6 & 2.4 / 5.4 \\
 & 0.95 & 83.3 & 98.8 & 97.4 & 1.2 / 2.6 \\
\bottomrule
\end{tabular}
\end{table}

Table~\ref{tab:uncertainty_results} presents the comprehensive results of our uncertainty-aware rejection analysis. At the baseline threshold ($\tau=0.50$), both models process 100\% of samples. The Teacher model achieves 97.4\% accuracy on clean data and 94.6\% under domain shift, corresponding to a degradation of 2.8\%. Student1 achieves 94.5\% accuracy on clean data and 84.4\% under shift, showing a 10.1\% degradation. This baseline comparison confirms that the Teacher model exhibits superior robustness to distribution shift.

As we increase the confidence threshold, both models demonstrate the expected trade-off: coverage decreases while accuracy on processed samples increases. At the intermediate threshold ($\tau=0.85$), the Teacher processes 94.6\% of samples with 98.9\% accuracy on clean data and 97.9\% accuracy under shift. The false-route rate (percentage of misclassified samples among processed ones) increases from 1.1\% on clean data to 2.1\% under shift. Student1 at this threshold processes 90.2\% of samples with 97.6\% clean accuracy and 94.6\% shifted accuracy, with false-route rates of 2.4\% and 5.4\% respectively.

In conservative mode ($\tau=0.95$), the Teacher maintains 99.5\% accuracy on clean data and 99.0\% accuracy under domain shift while processing 89.4\% of samples. The false-route rate remains below 1\% (0.5\% clean, 1.0\% shifted), making this operating point particularly suitable for clinical applications where sorting accuracy is paramount. Student1 achieves 98.8\% accuracy on clean data and 97.4\% under shift at 83.3\% coverage, with false-route rates of 1.2\% and 2.6\% respectively. These results demonstrate that both models can operate in a mode where ambiguous cells are rejected, ensuring that over 98\% of processed cells are correctly classified even under domain shift conditions.

\begin{figure}[ht]
\centering
\includegraphics[width=0.7\linewidth]{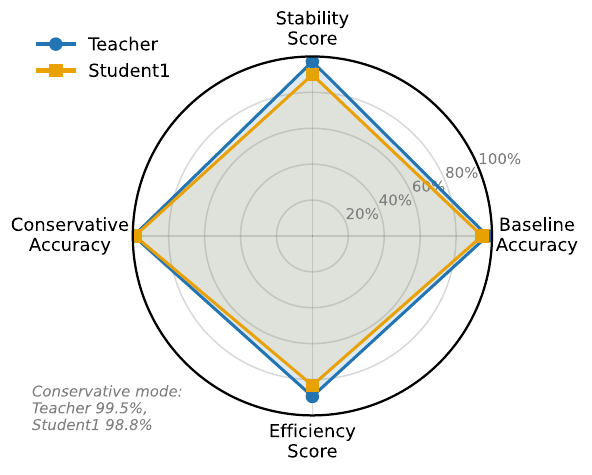}
\caption{Radar chart visualization of model robustness across four dimensions: Baseline Accuracy (performance at $\tau=0.50$), Stability Score (robustness to domain shift), Conservative Accuracy (performance at $\tau=0.95$), and Efficiency Score (coverage at $\tau=0.95$). The Teacher model (blue) demonstrates superior robustness to domain shift and maintains higher throughput in conservative mode, while Student1 (orange) achieves remarkable parameter efficiency with acceptable performance degradation under distribution shift.}
\label{fig:robustness_radar}
\end{figure}

Figure~\ref{fig:robustness_radar} synthesizes these findings in a radar chart visualization across four key dimensions: Baseline Accuracy (performance at $\tau=0.50$ with 100\% coverage), Stability Score (100\% minus degradation percentage, quantifying robustness to domain shift), Conservative Accuracy (performance at $\tau=0.95$ in high-confidence mode), and Efficiency Score (coverage maintained at $\tau=0.95$). This multi-dimensional view clearly illustrates that while the Student model achieves remarkable parameter efficiency with a 5000-fold compression, the Teacher maintains superior robustness to domain shift and higher throughput in conservative mode.

From a hardware implementation perspective, the confidence threshold mechanism can be realized with minimal overhead. After the FPGA computes the softmax probabilities, a simple comparator operation checks whether the maximum probability exceeds the threshold $\tau$. This comparison adds negligible latency (less than 0.1 microseconds) and requires minimal additional logic resources, ensuring that uncertainty-aware rejection does not compromise the sub-25 microsecond total latency budget reported in our main results. The threshold value can be configured as a runtime parameter, allowing system operators to adjust the throughput-safety trade-off based on specific application requirements.

These results establish a quantitative framework for deploying our models in clinical settings where both throughput and safety must be carefully balanced. Research applications prioritizing maximum sample throughput might operate at $\tau=0.85$ (over 90\% coverage with approximately 98\% accuracy), while clinical diagnostic applications demanding highest accuracy would operate at $\tau=0.95$ (over 83\% coverage with over 98\% accuracy). The explicit quantification of false-route rates under domain shift provides critical information for regulatory compliance and risk assessment in clinical deployment scenarios.


\end{document}